\documentclass{article}
\usepackage{cplus_preprint,times}

\usepackage{amsmath,amsfonts,bm}

\def\eqref#1{equation~\ref{#1}}

\def\1{\bm{1}}

\DeclareMathAlphabet{\mathsfit}{\encodingdefault}{\sfdefault}{m}{sl}
\SetMathAlphabet{\mathsfit}{bold}{\encodingdefault}{\sfdefault}{bx}{n}

\usepackage{amsmath,amssymb}
\usepackage{booktabs}
\usepackage{capt-of}
\usepackage{graphicx}
\usepackage{array}
\usepackage{tabularx}
\usepackage{enumitem}
\usepackage{placeins}
\usepackage{wrapfig}
\usepackage{algorithm}
\usepackage{algpseudocode}
\usepackage{microtype}
\usepackage{hyperref}
\usepackage[capitalise,noabbrev]{cleveref}
\usepackage{url}
\usepackage{fontawesome5}
\usepackage{comment}
\hypersetup{
    hidelinks,
    pdftitle={Continual Learning via Self-Probe Gradients},
    pdfauthor={Dongkyu Cho, Rumi Chunara, Sungmin Cha}
}

\newcommand\csm[1]{\textcolor{black}{#1}}

\title{Continual Learning via Self-Probe Gradients}

\author{Dongkyu Cho\\
New York University \\
\textit{dongkyu.cho@nyu.edu}
\And
Rumi Chunara\\
New York University \\
\textit{rumi.chunara@nyu.edu}
\And
Sungmin Cha\thanks{Corresponding author. This work, including all code, was conducted by the authors in their personal and academic capacity (co-authors at New York University) and does not represent the views of Meta.}\\
Meta Reality Labs \\
\textit{sungmincha@meta.com}}

\preprintfinalcopy

\begin{document}
\raggedbottom

\maketitle

\begin{abstract}

Adapting pretrained models to new data can cause catastrophic forgetting of previously learned behavior. When only a few past samples remain, they give continual learning methods sparse and narrow evidence about what to preserve. 
We show that language models can expand this evidence through self-probing, in which the frozen model generates new inputs from the retained samples and records its own predictions on them.
Unlike prior work that replays such data as training examples, our method, CPLUS uses self-probe and past-sample gradients to scale down parameter updates that conflict with prior behavior.
Experiments with five language models on four benchmarks show three results. First, the same probes preserve more prior behavior as gradient signals than as replay data. Second, CPLUS learns the new data while consistently reducing forgetting more than existing baselines, especially when past data are scarce, and this protection extends to benchmarks not used for training. Third, we observe that CPLUS also becomes more effective as models grow: within the Qwen3 model family, it recovers an increasing share of the forgetting caused by standard fine-tuning.

\end{abstract}

\begin{center}
\small
\href{https://github.com/umamicode/cplus-continual}
{\faGithub\quad Code: \nolinkurl{github.com/umamicode/cplus-continual}}
\end{center}

\section{Introduction}
\label{sec:introduction}

\begin{figure}[t]
    \centering
    \includegraphics[width=\linewidth]{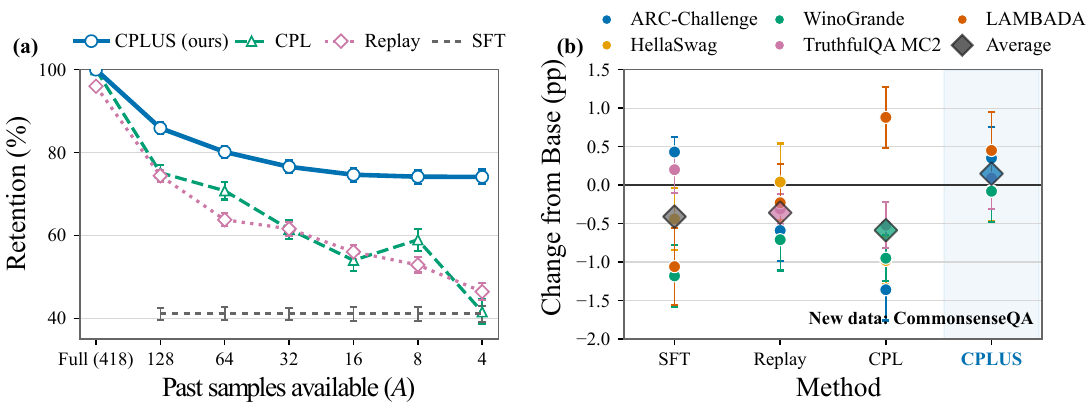}
    \caption{\textbf{Limited past-data access leaves preservation signals
    both sparse and narrow.}
    \textcolor[HTML]{0072B2}{\textbf{(a) Preservation weakens when past
    samples are scarce.}} Qwen3-0.6B is adapted to new CommonsenseQA data.
Replay, Collaborative Parameter Learning (CPL), and CPLUS (Ours) use
\(A\) retained examples (anchors); supervised fine-tuning (SFT)
uses only the new data. Retention is the percentage of initially correct examples that remain correct after adaptation (Correct \(\rightarrow\) Correct). It generally decreases as \(A\) shrinks, indicating increased forgetting. CPLUS achieves higher Retention across \csm{anchor} budgets.
    \textcolor[HTML]{D55E00}{\textbf{(b) Preserving the anchored target task does not necessarily preserve performance elsewhere.}}
    After the same CommonsenseQA adaptation, we evaluate five benchmarks that are not used for adaptation or as anchors. Colored points report each task's accuracy change from the base model, and diamonds report the five-task average. Points and error bars are means and standard errors over five seeds; all methods attain 100\% Acquisition (Incorrect
    \(\rightarrow\) Correct) on CommonsenseQA. CPLUS remains closest to the
    base model across the profile.}
    \label{fig:problem-motivation}
\end{figure}

\csm{Foundation models are pretrained with substantial compute, data, and engineering effort, yet often require further training to correct errors, incorporate new facts, or acquire domain-specific capabilities.}
Such updates can overwrite prior behavior, a phenomenon known as \csm{catastrophic forgetting~\citep{mccloskey1989catastrophic}.
Continual learning (CL) methods mitigate forgetting in several ways, and many of them rely on evidence of prior behavior, such as retained data or statistics derived from it~\citep{rolnick2019experiencereplaycontinuallearning,zhou2024continuallearningpretrainedmodels}}
For publicly released pretrained models\csm{~\citep{yang2025qwen3technicalreport, gemmateam2024gemma2improvingopen, abdin2024phi3technicalreporthighly}}, however, users typically receive model weights without the original training corpus or train-time statistics.
\csm{At best, they can collect a few examples of behavior worth keeping, such as inputs the model already answers correctly (\textit{i.e.}, anchors). Anchors give only a sparse and narrow view of the behavior that should be preserved, which makes forgetting harder to prevent~\citep{knoblauch2020optimalcontinuallearningperfect}. Figure \ref{fig:problem-motivation} shows both effects: preservation weakens as fewer anchors are available, and preserving the task they represent does not guarantee preservation elsewhere.}

A natural way to compensate for limited past data is to recover additional
evidence from the base model itself. 
\csm{Given a few past samples as in-context demonstrations~\citep{brown2020languagemodelsfewshotlearners}, the model can generate new inputs for the same task and label them with its own predictions, denoted as self-probes. Prior work has mainly used such model-generated data as reply data or distillation targets~\citep{huang2024mitigatingcatastrophicforgettinglarge}.}
\csm{When used as training targets, however, self-probes enter the update itself: possibly incorrect predictions are learned as if they were labels, and they do not indicate which parts of a new data update would disrupt prior behavior.}
In a complementary direction, \citet{yang2026cpl} show that, when the relevant past data are fully available, their gradients can identify \emph{collaborative parameters}, whose proposed updates are predicted to preserve prior behavior.
\csm{With only a few anchors, however, this evidence becomes sparse and narrow, and CPL forgets more (Figure \ref{fig:problem-motivation}). Together, these observations suggest using self-probe gradients to broaden the evidence for identifying collaborative parameters.}

We develop this idea as \textbf{CPLUS}: \textbf{C}ollaborative
\textbf{P}arameter \textbf{L}earning \textbf{U}sing \textbf{S}elf-Probes.
\csm{CPLUS combines gradients from anchors and self-probes to estimate, for each parameter, whether the update proposed by learning new data would preserve or disrupt prior behavior. It keeps collaborative updates and scales down those predicted to cause forgetting. Because these decisions can change from step to step, CPLUS smooths them over training steps into a soft mask. Note that the self-probes are never used as training targets. They only decide how much of each proposed update is applied, so an incorrect probe can slow down a change but cannot pull the model toward its own error.}

\csm{We evaluate CPLUS on three Qwen3 models (0.6B, 1.7B, and 4B), Gemma-2-2B-it, and Phi-3.5 Mini, using CommonsenseQA, ARC-Challenge, MMLU, and MedQA as new data with 4 to 128 anchors. CPLUS learns the new data while achieving higher Retention, the share of initially correct answers that remain correct, than all main baselines in 118 of 120 settings. The advantage is clearest when anchors are scarce: with four or eight anchors, CPLUS leads in all 24 Qwen 3 settings, by 7.73 points on average. 
It also preserves performance more broadly, maintaining near-base or better scores across five benchmarks not used for training (\Cref{fig:problem-motivation}(b)).
In sequential adaptation, CPLUS achieves 78.87\% average task accuracy, compared with 69.73\% for the strongest baseline.
In addition, CPLUS becomes more effective as models grow: under limited anchor budgets, the share of SFT forgetting it recovers rises from 46.6\% for Qwen3-0.6B to 69.4\% for Qwen3-4B.
Our contributions are as follows:}
\begin{itemize}[leftmargin=*,labelsep=0.45em,itemsep=1pt,topsep=2pt]
    \item We show, in matched comparisons, that self-probes preserve more prior behavior when used to guide updates than when replayed as training samples.
    \item We introduce CPLUS, which combines gradients from past samples and self-probes with a smoothed update mask to reduce forgetting under limited past-data access.
    \item \csm{We provide a systematic evaluation across five models, four benchmarks, and six anchor budgets, complemented by broader-capability, sequential, and compute studies.}
\end{itemize}

\section{Related Work}
\label{sec:related-work}

\paragraph{Continual Learning.}
Continual learning methods are commonly grouped into regularization-based
approaches, which discourage changes to parameters important for earlier
tasks \citep{kirkpatrick2017ewc,aljundi2018memoryawaresynapses, cha2021cpr};
rehearsal-based approaches, which store and replay past samples
\citep{rolnick2019experiencereplaycontinuallearning,rebuffi2017icarl, 10204252}; \csm{optimization-based approaches, which modify the update using gradients from past data~\citep{lopezpaz2017gem,chaudhry2019agem,farajtabar2020ogd}}; and
expansion-based approaches, which allocate additional capacity to balance
stability \& plasticity
\citep{yan2021der,wang2022foster,zhou2024continuallearningpretrainedmodels}.
These approaches navigate the trade-off between preserving prior behavior and learning new data~\csm{\citep{mccloskey1989catastrophic}}, commonly relying on past data or information derived from them \citep{cho2025forget,dohare2024plasticity}.
CPL~\citet{yang2026cpl} makes this dependence explicit, showing that
parameter-wise gradient analysis can retain updates predicted to learn the new data without interfering with prior behavior, approaching zero forgetting when past data are fully available.
Yet full past-data access is rarely possible when adapting open-weight
models, which are typically released without their original training data or associated statistics
\citep{zhou2024continuallearningpretrainedmodels,marek2026forgetting}.
\csm{Note that CPLUS follows this optimization-based line of work. Instead of projecting the whole update as A-GEM~\citep{chaudhry2019agem} and OGD~\citep{farajtabar2020ogd} do, it rescales each coordinate of the proposed update.}

\paragraph{Self-Generated Data.}
The model's own generations provide one source of this additional evidence,
but existing methods differ in how they use the generated data. Prior work
shows that replay can reduce forgetting and improve data efficiency during
adaptation when the replay data approximate the past-data distribution
\citep{bethune2025scalinglawsforgetting,kotha2026replayingpretrainingdata,
marek2026forgetting}.
Accordingly, generative replay and Self-Synthesized Rehearsal (SSR) seek to
approximate unavailable past data and train directly on generated samples
\citep{shin2017continuallearningdeepgenerative,
huang2024mitigatingcatastrophicforgettinglarge,
resta2024selfgeneratedreplaymemoriescontinual}.
Self-Distillation Fine-Tuning instead uses a demonstration-conditioned model
as an on-policy teacher \citep{shenfeld2026selfdistillation}.
CPLUS uses self-generated data differently. Rather than training on
self-probes as replay data, it uses their gradients to identify
which parts of a proposed new-data update may interfere with prior behavior.
Thus, new data determine what the model learns, while self-probes guide how it
learns without disrupting what it already knows.
See Appendix \ref{app:extended-related-work} for more.

\section{CPLUS: Motivation and Method}
\label{sec:cplus}

\csm{This section first shows why limited past-data access weakens preservation (\Cref{sec:problem}) and then introduces CPLUS (\Cref{sec:method}).}

\subsection{Continual Learning under Limited Past-Data Access}
\label{sec:problem}

We study continual learning when only \(A\) retained past samples, which we
call \textit{anchors}, remain available from the broader pool of past data.
We examine how both the amount and coverage of this evidence affect
the preservation of prior behavior.

The two limitations illustrated in \Cref{fig:problem-motivation}
motivate our method.
\textcolor[HTML]{0072B2}{\textbf{First,}} preservation weakens as fewer
anchors are available. 
\Cref{fig:problem-motivation}(a) holds the \csm{anchored} task fixed and reduces the CommonsenseQA anchors from 418 to four; Retention decreases for both replay and CPL as \(A\) shrinks.
\textcolor[HTML]{D55E00}{\textbf{Second,}} the available anchors represent
only a narrow part of a base model's capabilities.
\Cref{fig:problem-motivation}(b) evaluates five benchmarks used neither for
adaptation nor as anchors and shows that preserving the represented target
does not necessarily preserve behavior beyond it.

\csm{Both limitations arise because all preservation evidence comes from the anchors.}
\csm{When this evidence is sparse, it is hard to tell which changes will preserve prior behavior. When it is narrow, preserving the anchored task does not guarantee preserving other capabilities. The challenge, therefore, is to obtain a broader preservation signal despite limited access to past data.}

\csm{One way to broaden this signal is to generate self-probes from the base model. Replaying them as training data~\citep{huang2024mitigatingcatastrophicforgettinglarge}, treats the model's predictions as labels even though they may be wrong. Learning from incorrect predictions can reinforce errors through confirmation bias~\citep{arazo2020pseudolabeling}, and noisy replay labels can exacerbate forgetting~\citep{kim2021continuallearningnoisydata}. We instead treat self-probes as records of the frozen base model's behavior and use their gradients only to assess proposed new data updates.}

\FloatBarrier

\subsection{Proposed Method}
\label{sec:method}

\paragraph{Notation and Preliminaries.}
Building on the parameter-wise view of \csm{CPL}~\citep{yang2026cpl}, we let new
data propose updates and use anchor and self-probe gradients to assess
which changes may disrupt prior behavior.
Let \(f_{\theta_0}\) denote the frozen pretrained base model, and let
\(\theta_t\) denote the adapted parameters before update \(t\geq1\),
initialized as \(\theta_1=\theta_0\). 
Adaptation learns from \(\mathcal D_{\mathrm{new}}\), with only \(A\) retained past samples in the anchor set \(\mathcal P_A\). Let \(\delta_t\) denote the optimizer's proposed new-data update at adaptation step \(t\), and let \(g_A\) denote the most recently computed mean anchor-loss gradient. The product
\(g_{A,i}\delta_{t,i}\) estimates parameter \(i\)'s first-order contribution
to the change in anchor loss. We call \csm{parameter $i$}
\emph{collaborative} when this product is nonpositive and
\emph{conflicting} otherwise. These labels depend on the current update and
preservation signal; they are not fixed parameter types.

\begin{figure*}[t]
    \centering
    \includegraphics[width=\textwidth]{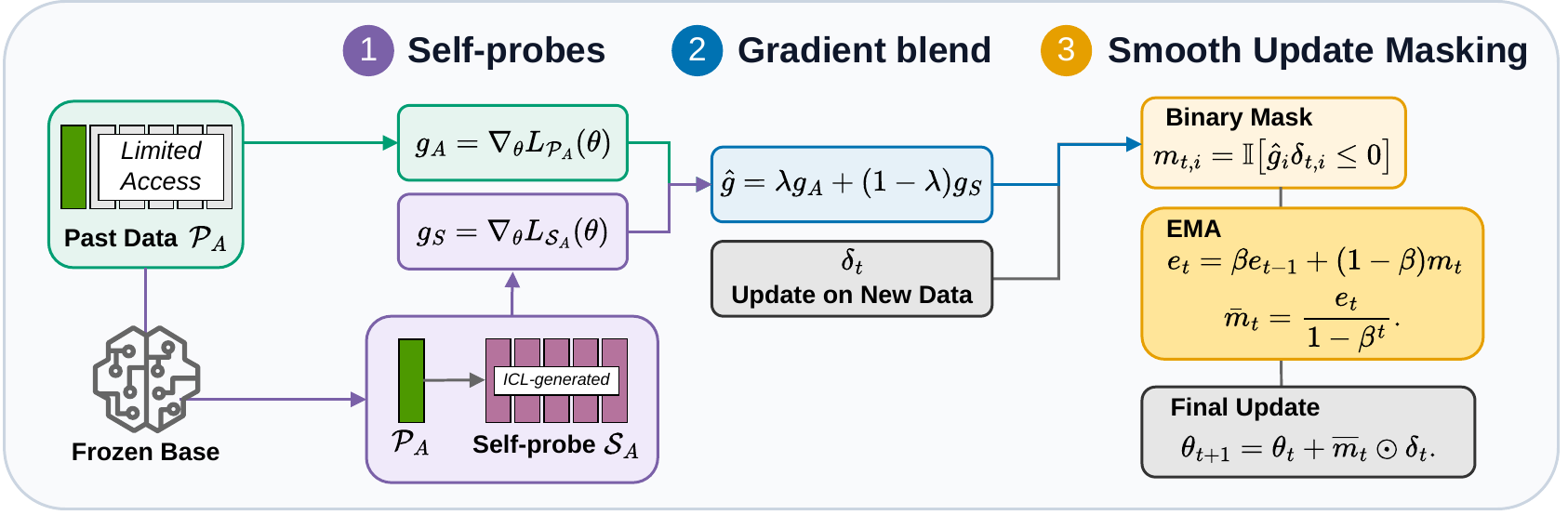}
    \caption{\textbf{Overview of CPLUS.}
    \textcolor[HTML]{7554B3}{\textbf{1. Self-probes.}}
    The frozen base model uses the limited anchors \(\mathcal P_A\) as
    in-context demonstrations to generate a self-probe bank \(\mathcal S_A\).
    \textcolor[HTML]{0072B2}{\textbf{2. Gradient blend.}}
    Gradients from the anchors and self-probes are blended into \(\widehat g\).
    \textcolor[HTML]{E69F00}{\textbf{3. Smooth Update Masking.}}
    The blended preservation gradient gates the optimizer's proposed new-data
    update \(\delta_t\) \csm{parameter}-wise, and the binary mask is smoothed by an
    exponential moving average. Finally, the smoothed mask is applied to
    \(\delta_t\). Thus, self-probes supply preservation evidence rather than
    replay targets.}
    \label{fig:method-overview}
\end{figure*}

To broaden the evidence used by this criterion, we propose \textbf{CPLUS}:
Collaborative Parameter Learning Using Self-Probes.
CPLUS generates self-probes from the frozen base model and blends their
gradients with anchor gradients. It then uses this augmented
signal to identify collaborative \csm{parameters} and smooths the resulting
decisions over time. \Cref{fig:method-overview} summarizes these three
components.

\subsubsection{Self-Probes}
\label{sec:self-probes}

Before adaptation, CPLUS uses the frozen base model \(f_{\theta_0}\) to
generate self-probes that reflect its prior behavior. For each probe, a small
context \(C_j\subseteq\mathcal P_A\) provides in-context demonstrations
\citep{brown2020languagemodelsfewshotlearners}. The model generates a new
input \(\widetilde x_j\) and assigns a prediction \(\widetilde y_j\) using a
fixed decoding rule. Repeating this process produces a self-probe bank
\(\mathcal S_A=\{(\widetilde x_j,\widetilde y_j)\}_{j=1}^{N}\), which
remains fixed during adaptation. The self-probes record the base model's
predictions rather than externally verified labels.

During adaptation, the fixed self-probe bank supplies gradients that guide
proposed new-data updates. Neither the anchor loss nor the self-probe loss is added to the
new-data objective that produces the proposed update \(\delta_t\). Their
gradients determine how much of that proposal is applied to each parameter.
We compare this use of self-probes with direct replay in
\Cref{fig:self-probe-role-anchor-scale}; construction and audit details
appear in \Cref{app:method-details,app:self-probe-audit}.

\subsubsection{Gradient Blending}
\label{sec:gradient-blending}

To combine evidence from the two sources, CPLUS computes mean anchor and
self-probe gradients at the beginning of each epoch, using the current
adapted model parameters \(\theta\):
\begin{equation}
    g_A=\nabla_\theta L_{\mathcal P_A}(\theta),
    \qquad
    g_S=\nabla_\theta L_{\mathcal S_A}(\theta).
\end{equation}
It combines them into a preservation gradient,
\begin{equation}
    \widehat g=\lambda g_A+(1-\lambda)g_S.
\end{equation}
We use \(\lambda=1/2\) and do not normalize or rescale either gradient. The
two sources therefore receive equal coefficients, but their influence on any
particular parameter can differ with gradient magnitude. We reuse \(g_A\), \(g_S\), and \(\widehat g\) throughout each epoch,
while computing the new update \(\delta_t\) at every step.

This blend provides one preservation signal that accounts for evidence from
both sources. By linearity, \(\widehat g\) is the gradient of
\(\lambda L_{\mathcal P_A}+(1-\lambda)L_{\mathcal S_A}\), evaluated at the
refresh parameters \(\theta\). This weighted preservation loss supplies the
signal for masking proposed updates; it is not added to the new-data
training objective.
\Cref{tab:cplus-ablation} compares alternative ways to construct the
gradients.

\subsubsection{Smooth Update Masking}
\label{sec:update-masking}

At each adaptation step, CPLUS uses the most recently refreshed
\(\widehat g\) to assess the optimizer's proposed new-data update
\(\delta_t\). For parameter \(i\), it forms the binary mask
\begin{equation}
    m_{t,i}
    =
    \mathbb I\!\left[\widehat g_i\delta_{t,i}\leq 0\right].
    \label{eq:mask}
\end{equation}
Here, \(m_{t,i}=1\) identifies a \csm{parameter} as collaborative with respect to
the blended preservation \csm{gradient}, while \(m_{t,i}=0\) identifies conflict.
The criterion uses the optimizer's proposed parameter change, not the raw
new-data gradient.

\csm{Because $\delta_t$ is recomputed at each step while $\widehat{g}$ is refreshed only once per epoch, a parameter's mask can flip from step to step.}
CPLUS smooths them with a bias-corrected exponential moving
average:
\begin{equation}
    e_t=\beta e_{t-1}+(1-\beta)m_t,
    \qquad
    \overline m_t=\frac{e_t}{1-\beta^t},
\end{equation}
Here \(m_t\) is the binary mask from \csm{\Cref{eq:mask} and}
\(\overline m_t\) is its bias-corrected, smoothed version. We set
\(e_0=0\) and \(\beta=0.9\). The applied update is:
\begin{equation}
    \theta_{t+1}
    =
    \theta_t+\overline m_t\odot\delta_t.
\end{equation}
Intuitively, persistent collaborative decisions move a mask coefficient
toward one, while persistent conflicting decisions move it toward zero.
Because the coefficients lie in \([0,1]\), the mask can limit each proposed
update but cannot amplify or reverse it. Smoothing \csm{also} relaxes the instantaneous sign constraint in exchange
for less sensitivity to isolated mask flips. Our ablations show that it improves Retention
at full Acquisition (\Cref{tab:cplus-ablation,tab:ema-selection}).
Implementation details appear in \Cref{app:method-details}.

\section{Experiments}
\label{sec:experiments}

We evaluate CPLUS under limited data access and beyond the anchored task,
compare ways of using self-probes, and then study sequential learning,
training cost, and learning dynamics.

\subsection{Experimental Setting}
\label{sec:experimental-setting}

\textbf{Models and benchmarks.}
We study within-family scale with Qwen3-0.6B, 1.7B, and 4B
\citep{yang2025qwen3technicalreport}, and test out-of-family generalization
with Gemma-2-2B-it \citep{gemmateam2024gemma2improvingopen} and the Phi-3.5
Mini model \citep{abdin2024phi3technicalreporthighly}.
The adaptation benchmarks are CommonsenseQA
\citep{talmor2019commonsenseqa}, ARC-Challenge \citep{clark2018arc}, MMLU
\citep{hendrycks2021mmlu}, and MedQA \citep{jin2021medqa}.
Following \citet{yang2026cpl}, we train on initially incorrect examples
and measure forgetting on initially correct examples, from which anchors
are drawn. This lets us measure what the model learns and what it
forgets separately.

We vary the anchor budget over \(A\in\{4,8,16,32,64,128\}\); plots show the
count, whereas tables also report its percentage of the full past dataset.
Baselines are standard supervised fine-tuning (SFT), exemplar replay
\citep{rolnick2019experiencereplaycontinuallearning}, self-probe replay
\citep{resta2024selfgeneratedreplaymemoriescontinual}, and CPL
\citep{yang2026cpl} under matched data access. Frozen distillation
\citep{hinton2015distilling}, Self-Synthesized Rehearsal (SSR)
\citep{huang2024mitigatingcatastrophicforgettinglarge}, A-GEM
\citep{chaudhry2019agem}, and OGD-GTL \citep{farajtabar2020ogd} appear only
in the compute comparison (\Cref{fig:retention-compute}). Ablations appear in \Cref{tab:cplus-ablation}.

\textbf{Hyperparameter protocol.}
Following the two-phase protocol of \citet{cha2025hyperparameters}, we use
Qwen3-0.6B on CommonsenseQA for development. We retain CPL's 25-epoch
training budget \citep{yang2026cpl} and use \texttt{lr=1e-6} by default.
The CPLUS-specific defaults are \(N=512\) self-probes, two in-context
examples, generation temperature \(0.8\), equal-weight mean-gradient
blending, no norm alignment, and mask EMA \(\beta=0.9\).
We keep the self-probe and masking settings fixed across models, benchmarks,
and anchor budgets. The
compute comparison uses 32 self-probes; other main comparisons use
the default bank.

\begin{figure}[!tp]
    \centering
    \includegraphics[width=\textwidth]{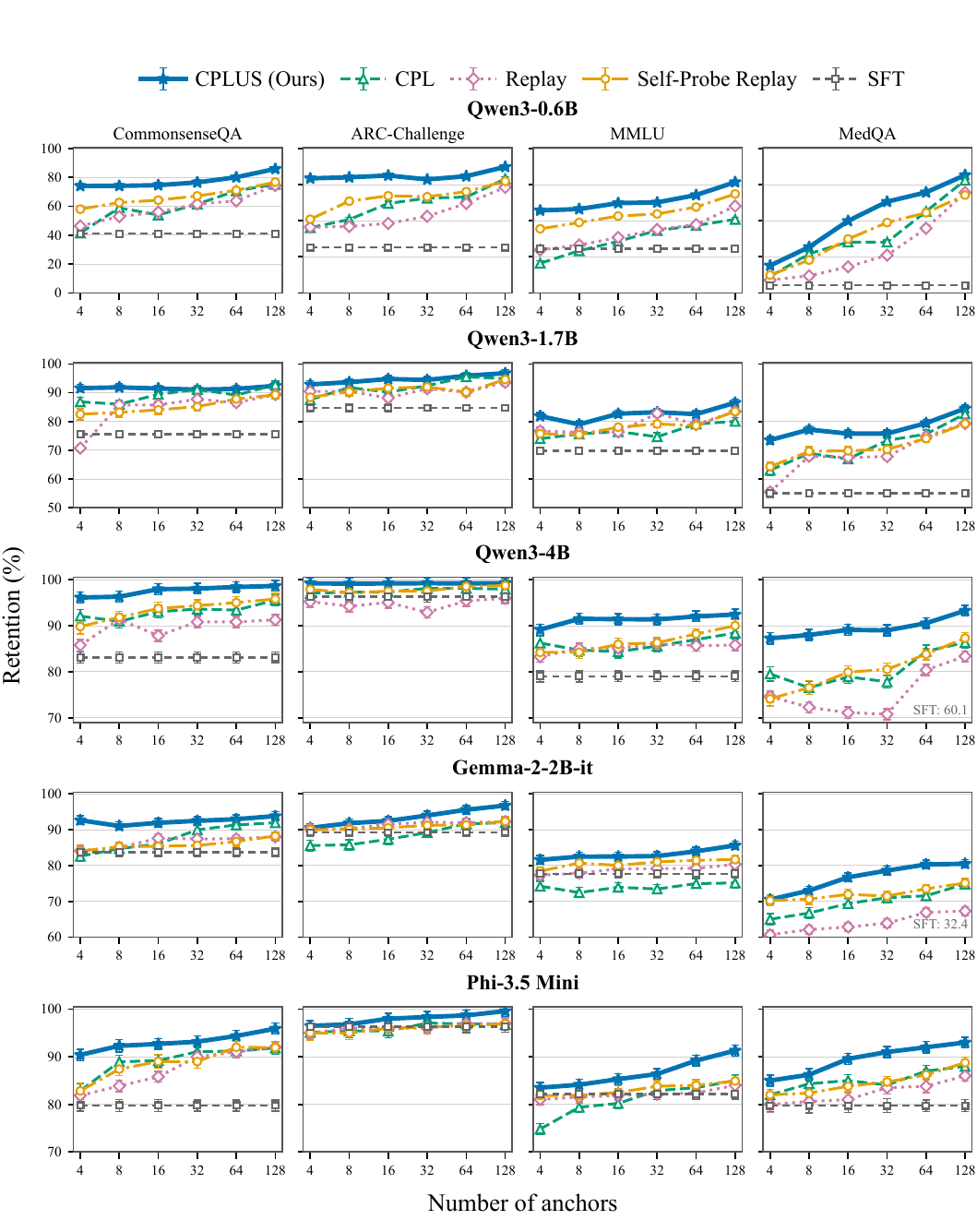}
    \caption{\textbf{CPLUS reduces forgetting even with few past examples.} Curves show \textcolor[HTML]{0072B2}{\textbf{CPLUS (Ours)}}, \textcolor[HTML]{009E73}{\textbf{CPL}}, \textcolor[HTML]{CC79A7}{\textbf{Replay}}, \textcolor[HTML]{E69F00}{\textbf{Self-Probe Replay}}, and \textcolor[HTML]{666666}{\textbf{SFT}} across anchor budgets. Retention is the percentage of initially correct examples that remain correct after adaptation (Correct \(\rightarrow\) Correct). Error bars show five-seed standard errors. Panels share a vertical scale within each model row. Off-scale MedQA SFT values are annotated in the 4B and Gemma rows.}
    \label{fig:main-results}
\end{figure}

\textbf{Evaluation.}
Retention (Correct \(\rightarrow\) Correct) is
\(\Pr(\text{correct after}\mid\text{correct before})\), and Acquisition
(Incorrect \(\rightarrow\) Correct) is
\(\Pr(\text{correct after}\mid\text{incorrect before})\). For sequential
adaptation, Average Task Accuracy is the unweighted mean accuracy over tasks
encountered so far. Primary single-task and broader-capability comparisons report
means and standard errors over five runs.

\subsection{Preservation with Limited Anchors}
\label{sec:experimental-results}

We first evaluate whether CPLUS reduces forgetting under limited data access.
In \Cref{fig:main-results}, CPLUS achieves higher Retention than the
strongest baseline in 70 of 72 Qwen3 settings, including all 24 at
\(A\in\{4,8\}\). At these budgets, its mean gain is 7.73 points,
while recovery of SFT forgetting rises from 46.6\% at 0.6B to 50.6\% at 1.7B
and 69.4\% at 4B (\Cref{fig:model-scale}). CPLUS also leads in all 48 Gemma
and Phi-3.5 Mini settings. 
All methods in the grid attain 100\% Acquisition, yet differ in Retention even when the adaptation examples are fully learned.

\subsection{Preservation Beyond the Represented Target}
\label{sec:broader-preservation}

\begin{figure}[!htbp]
    \centering
    \includegraphics[width=\textwidth]{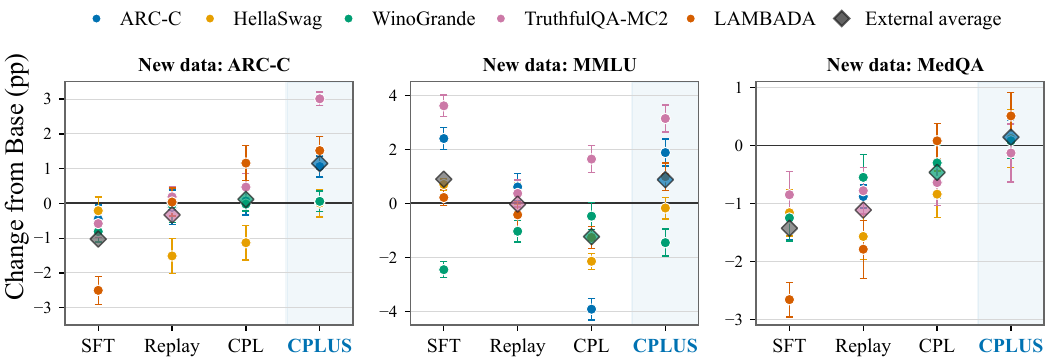}
    \caption{\textbf{CPLUS preserves broader performance on other tasks while learning new data.}
    Panels show changes from the base model after ARC-C, MMLU, or MedQA
    adaptation. Diamonds are unweighted averages over benchmarks external to
    the adaptation task; in the ARC-C panel, its own point is excluded from
    that average. Points and error bars show means and standard errors over
    five seeds. Each panel has its own labeled vertical range.}
    \label{fig:medqa-broader-capability}
\end{figure}
Retention on the anchored task does not establish preservation elsewhere.
After CommonsenseQA adaptation, the right panel of
\Cref{fig:problem-motivation} shows broader losses with CPL on four of five
benchmarks, while CPLUS stays near or above the base model on each.

The pattern extends to three other adaptation tasks
(\Cref{fig:medqa-broader-capability}). On benchmarks external to each new
task, CPLUS improves average performance relative to the base model by 1.15, 0.88, and 0.14 percentage points after ARC-C, MMLU, and MedQA adaptation, respectively.. It has the highest
external average after ARC-C and MedQA adaptation and is within 0.02 points
of SFT after MMLU adaptation.

We also test how anchor composition affects preservation and error correction
by varying whether anchors were initially answered correctly
(\Cref{app:anchor-composition,tab:anchor-composition}).

\begin{figure}[!htbp]
    \centering
    \begin{minipage}[t]{0.56\linewidth}
        \vspace{0pt}
        \centering
        \includegraphics[width=\linewidth]{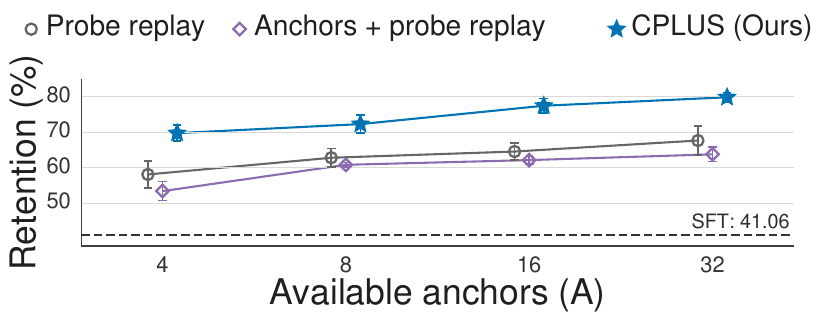}
        \caption{\textbf{Self-probes work better as gradients than replay.}
        Mean Retention and standard error over three matched seeds on
        Qwen3-0.6B/CommonsenseQA. Methods use the same probes and update
        budget.}
        \label{fig:self-probe-role-anchor-scale}
    \end{minipage}\hfill
    \begin{minipage}[t]{0.42\linewidth}
        \vspace{0pt}
        \centering
        \footnotesize
        \setlength{\tabcolsep}{3pt}
        \begin{tabular}{@{}lr@{}}
            \toprule
            Variant & Retention (\%) \\
            \midrule
            Anchor gradient only & 55.13 \\
            Probe gradient only & 71.49 \\
            CPLUS without EMA & 78.71 \\
            \textbf{CPLUS} & \textbf{79.96} \\
            CPLUS + norm alignment & 79.94 \\
            \bottomrule
        \end{tabular}
        \vspace{1.2em}
        \captionof{table}{\textbf{Component ablation.} Qwen3-0.6B,
        CommonsenseQA, \(A=8\); one seed. All variants attain 100\%
        Acquisition.}
        \label{tab:cplus-ablation}
    \end{minipage}
\end{figure}

\subsection{How Should Self-Probes Be Used?}
\label{sec:fusion-analysis}

We compare alternative uses of the same self-probes. Across four anchor budgets, CPLUS retains more previously correct
answers than either replay variant (\Cref{fig:self-probe-role-anchor-scale}).
Here, Probe replay trains on frozen-model
labels, while anchor + probe replay additionally includes the available real
anchors. All methods attain near-perfect Acquisition.

We next isolate the contributions of anchors, self-probes, and smoothing
(\Cref{tab:cplus-ablation}). Combining both gradient sources performs best
in this diagnostic; smoothing adds a smaller gain, while norm alignment
changes little.

We further test whether smaller updates alone explain the gains.
Neither matched update-norm nor randomized-mask controls reproduce CPLUS's
Retention at full Acquisition, suggesting that coordinate selection matters
beyond aggregate attenuation in this diagnostic
(\Cref{tab:matched-update-controls,app:gradient-source-dynamics}).

\subsection{Sequential Multi-Task Adaptation}
\label{sec:sequential-results}

\begin{figure}[!htbp]
    \centering
    \includegraphics[width=\textwidth]{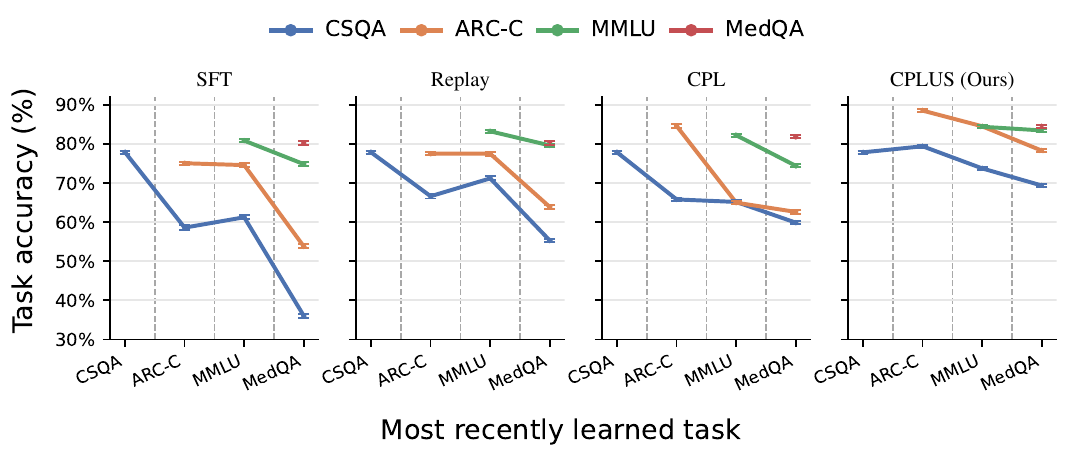}
    \caption{\textbf{CPLUS in limited-anchor sequential multi-task
    adaptation.} Training follows CSQA \(\to\) ARC-C \(\to\) MMLU \(\to\)
    MedQA, carrying forward each checkpoint; MMLU thus inherits CSQA and ARC-C
    training. The CSQA checkpoint is shared. Curves show accuracy on all
    encountered tasks (mean and standard error, three seeds). Replay, CPL,
    and CPLUS use \(A=8\) anchors per past task.}
    \label{fig:sequential-multitask}
\end{figure}

We next test whether these benefits persist when learning multiple tasks
sequentially. We adapt to CommonsenseQA, ARC-Challenge, MMLU, and MedQA in
that order, with eight anchors per preceding task. After the final stage,
CPLUS achieves 78.87\% Average Task Accuracy, compared with 69.73\% for
replay and 69.65\% for CPL; it also has the highest accuracy on every task
in this sequence (\Cref{fig:sequential-multitask}). Full results appear in
\Cref{tab:sequential-stage-summary,tab:sequential-final}.

\subsection{Efficiency and Optimization Robustness}
\label{sec:efficiency-robustness}

\begin{figure}[t]
    \centering
    \includegraphics[width=\textwidth]{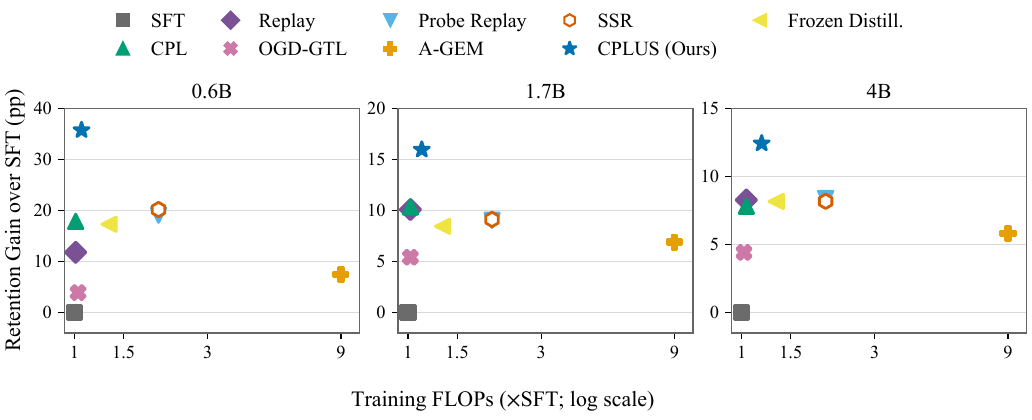}
    \caption{\textbf{CPLUS offers a favorable compute--Retention tradeoff across the evaluated model scales.}
    On CommonsenseQA, the axes report Retention gain over SFT and
training FLOPs relative to SFT (log scale). Preservation methods use
    eight anchors; SFT uses none.
    One-time preparation is excluded. App. \ref{app:recurring-compute} gives the
    accounting and baseline definitions. Y-axis scales differ.}
    \label{fig:retention-compute}
\end{figure}

\paragraph{Training cost.}
We examine the compute--Retention tradeoff using a smaller self-probe bank.
\Cref{fig:retention-compute} compares Retention gain against training
FLOPs. Across the displayed model sizes, CPLUS achieves a larger
gain than the other evaluated methods.

\begin{figure}[!htbp]
    \centering
    \includegraphics[width=\textwidth]{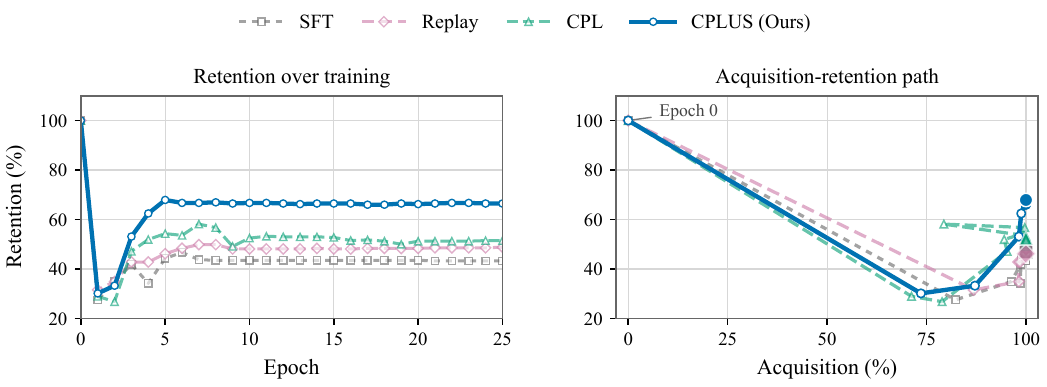}
    \caption{\textbf{Early forgetting can be followed by recovery as training continues.}
    Retention over epochs (left) and the Acquisition--Retention path (right)
    for a 25-epoch Qwen3-0.6B/CommonsenseQA diagnostic with eight anchors. The lines show epoch-level
    scores without uncertainty bands. This diagnostic is independent of the
    five-run comparison in \Cref{fig:main-results}.
    Filled markers indicate the first epoch reaching 100\% Acquisition.}
    \label{fig:learning-dynamics}
\end{figure}

\paragraph{Learning dynamics.}
Finally, we examine how learning and forgetting evolve during adaptation.
In an independent limited-anchor diagnostic, CPLUS first reaches 100\%
Acquisition after five epochs, compared with ten for CPL, and thereafter
maintains the highest Retention (\Cref{fig:learning-dynamics}). Replay initially
forgets less than SFT and remains above it after 25 epochs, but trails CPLUS
in Retention even though both reach full Acquisition by epoch five. Its gap
therefore reflects weaker preservation, not slower acquisition. The
trajectories also reveal a sharp early drop in Retention followed by partial
recovery; \Cref{app:epoch-level-dynamics} examines this pattern in more detail.
\Cref{app:fixed-aggressive-update} tests robustness under a larger-learning-rate
schedule. These trajectories distinguish temporary forgetting from the loss that remains after adaptation.

\FloatBarrier
\section{Conclusion}
\label{sec:conclusion}

We introduced CPLUS, a continual-learning method for adapting foundation
models when only a limited anchor set is available. CPLUS elicits self-probes
from a frozen base model and uses their gradients, together with anchor
gradients, as preservation signals that softly constrain updates on new data.
Across model sizes, benchmarks, and anchor budgets, CPLUS substantially
improves Retention while maintaining Acquisition, including beyond the task
represented by the anchors and in a tested four-task sequence. These results
show that self-probes need not be replayed as training examples: they can
instead provide actionable evidence for preserving prior behavior during
continual adaptation under limited access to past data.

\clearpage
\section*{Acknowledgments}
We thank NYU HPC for their generous support and assistance with computational
simulations and experiments. We are also grateful for the support from the 2026
Google Research Awards and Google's
\href{https://sites.research.google/trc/}{TPU Research Cloud (TRC) program} in their TPU compute grant. 

\bibliography{references}
\bibliographystyle{cplus}

\clearpage
\appendix
\setlength{\intextsep}{10pt plus 2pt minus 2pt}
\setlength{\textfloatsep}{12pt plus 2pt minus 2pt}
\setlength{\floatsep}{10pt plus 2pt minus 2pt}

\section{Experiment Index}
\label{app:experiment-index}

\begin{table}[H]
    \centering
    \caption{\textbf{Experiment index.}}
    \label{tab:experiment-index}
    \footnotesize
    \setlength{\tabcolsep}{4pt}
    \renewcommand{\arraystretch}{1.12}
    \begin{tabularx}{\textwidth}{@{}p{0.07\textwidth}X>{\raggedright\arraybackslash}p{0.34\textwidth}@{}}
        \toprule
        \textbf{ID} & \textbf{Question} & \textbf{Location} \\
        \midrule
        E1 & Does CPLUS reduce forgetting when anchors are limited? &
        \Cref{fig:main-results,app:model-scale} \\
        E2 & Does preservation extend beyond the behavior represented by the anchors? &
        \Cref{fig:medqa-broader-capability,app:broader-capability} \\
        E3 & Are self-probes more effective as preservation gradients than as replay targets? &
        \Cref{fig:self-probe-role-anchor-scale} \\
        E4 & Which CPLUS components and configurations matter? &
        \Cref{tab:cplus-ablation,app:hyperparameter-selection} \\
        E5 & Does CPLUS remain effective in the tested task sequence? &
        \Cref{fig:sequential-multitask,app:sequential-multitask} \\
        E6 & How do Acquisition, Retention, and computational cost evolve during adaptation? &
        \Cref{fig:retention-compute,app:epoch-level-dynamics} \\
        E7 & How does anchor composition affect preservation? &
        \Cref{app:anchor-composition} \\
        E8 & Does CPLUS remain effective under an aggressive update schedule? &
        \Cref{app:fixed-aggressive-update} \\
        E9 & Does update location matter beyond update magnitude or mask density? &
        \Cref{app:matched-update-controls,app:gradient-source-dynamics} \\
        E10 & How does the learned mask relate to full-data and layer-wise references? &
        \Cref{app:mask-behavior,app:layerwise-mask-audit} \\
        \bottomrule
    \end{tabularx}
\end{table}

\section{Additional Problem-Formulation Details}
\label{app:problem-details}

Let \(\mathcal P\) denote the unavailable full set of past examples,
\(g_{\mathcal P,t}\) its loss gradient at adaptation step \(t\), and
\(\widehat g_t\) an estimate from available evidence. For the instantaneous
mask, before temporal smoothing, let
\[
\widehat m_{t,i}
=
\mathbb I[\widehat g_{t,i}\delta_{t,i}\leq0],
\qquad
\widehat\delta_t
=
\widehat m_t\odot\delta_t.
\]
The masked update satisfies
\(\widehat g_t^\top\widehat\delta_t\leq0\), but its first-order effect on
the full past-data loss obeys
\[
\begin{aligned}
g_{\mathcal P,t}^{\top}\widehat\delta_t
&=
\widehat g_t^{\top}\widehat\delta_t
+
(g_{\mathcal P,t}-\widehat g_t)^{\top}\widehat\delta_t\\
&\leq
\lVert g_{\mathcal P,t}-\widehat g_t\rVert_2
\lVert\widehat\delta_t\rVert_2.
\end{aligned}
\]
Thus, forgetting may arise from estimation error or from a large update in
directions where the preservation estimate is inaccurate. Shrinking the update
to zero is not a satisfactory solution because it also prevents acquisition.

Because the mask acts on the optimizer update, the relevant errors depend on
that proposed update. Define
\[
c_{t,i}=g_{\mathcal P,t,i}\delta_{t,i},
\qquad [z]_+=\max(z,0).
\]
An estimated mask admits harmful update mass when it retains a coordinate with
\(c_{t,i}>0\), and removes useful update mass when it suppresses a coordinate
with \(c_{t,i}<0\). We therefore use
\[
E_{\mathrm{harmful\text{-}admitted}}
=
\sum_i\widehat m_{t,i}[c_{t,i}]_+,
\qquad
E_{\mathrm{safe\text{-}removed}}
=
\sum_i(1-\widehat m_{t,i})[-c_{t,i}]_+.
\]
These diagnostics distinguish preservation obtained by selectively blocking
harmful updates from preservation obtained by suppressing most update energy.

\section{Detailed CPLUS Method}
\label{app:method-details}

This section expands the three components summarized in
\Cref{sec:method}. Throughout, let
\(f_{\theta_0}\) be the frozen base model, \(\mathcal D_{\mathrm{new}}\) the
new-data set, and \(\mathcal P_A\) the limited anchor set of size \(A\).

\subsection{Constructing the Self-Probe Bank}

For each probe request \(j\), we sample a context
\(C_j\subseteq\mathcal P_A\) containing \(K\) anchors. The frozen model
uses this context to approximate the task and generate a new input. Optionally,
generation can include a generic topic cue \(c_j\) drawn from 24 predefined
broad areas; the reported categorical runs used this option. The resulting
generation is
\[
\widetilde x_j
\sim
q_{\theta_0}(\cdot\mid C_j,c_j;T),
\]
where \(T\) is the generation temperature and \(c_j=\varnothing\) when no cue is
used. We then apply the fixed decoding
rule of the base model to obtain a behavioral target,
\[
\widetilde y_j=h_{\theta_0}(\widetilde x_j).
\]
The target therefore records how the base model responds; it is not an
externally verified label. Outputs that do not satisfy the required task
interface are discarded, and duplicate probe inputs are removed. Generation
continues until the bank contains \(N\) valid pairs,
\[
\mathcal S_A
=
\{(\widetilde x_j,\widetilde y_j)\}_{j=1}^{N}.
\]
In a one-seed diagnostic, including the topic cue did not consistently improve
Retention.

The same frozen model supplies both the in-context generation distribution and
the behavioral targets. The bank is constructed before adaptation, cached, and
held fixed throughout a run. Neither \(\mathcal P_A\) nor \(\mathcal S_A\) is
added to the adaptation objective. They are used only to estimate which
components of an adaptation update should be preserved. In the reported
configuration, \(K=2\), \(T=0.8\), and \(N=512\).

Self-probes are intended to broaden behavioral coverage, not to reconstruct the
unavailable past-data distribution. Their usefulness depends on what the
frozen model can infer from the context. Model scale may also affect baseline
susceptibility to forgetting; \Cref{app:model-scale} therefore reports both
absolute Retention and recovery relative to that baseline.

\subsection{Self-Probe Bank Audit}
\label{app:self-probe-audit}

We audit the sealed self-probe banks for duplication, model confidence, and
overlap with the evaluation questions. Across three banks, all 1,536
normalized probe prompts are unique. The frozen model's assigned behavioral
targets have mean, median, and tenth-percentile confidence of 0.915, 0.984,
and 0.696, respectively. A post-hoc comparison with 1,096 evaluation prompts
finds one normalized exact match (0.07\%) and two probes with token Jaccard
similarity of at least 0.9 (0.13\%). These targets are frozen-model behavioral
predictions rather than independently verified ground-truth annotations.

\subsection{Why Use Self-Probes as Preservation Evidence?}

The key design choice is how generated data enter adaptation. Direct
self-generated replay treats the model outputs as additional training targets.
For example, an augmented replay objective would form a gradient of the form
\[
g_{\mathrm{replay}}
=
g_{\mathrm{new}}
+
\alpha\bigl(\lambda g_A+(1-\lambda)g_S\bigr),
\]
where \(g_{\mathrm{new}}\) is the new-data gradient and \(\alpha\) controls
rehearsal. The probe gradient can therefore create an optimization direction
even when it is unrelated to acquiring the new behavior. Repeated replay may also
reinforce systematic errors inherited from the frozen model.

CPLUS instead computes the optimizer proposal \(\delta_{\mathrm{new}}\) using only the
supervised adaptation objective and uses the probes to control that proposal:
\[
\delta_{\mathrm{CPLUS}}
=
\overline m(g_A,g_S,\delta_{\mathrm{new}})\odot\delta_{\mathrm{new}},
\qquad
\overline m_i\in[0,1].
\]
Consequently, a self-probe cannot introduce, amplify, or reverse an update
coordinate; it can only attenuate a component of the supervised adaptation
update. This makes probe use conditional on the update that is actually being
proposed, rather than rehearsing every generated example indiscriminately. It
also keeps acquisition supervision separate from preservation evidence.

To test this distinction, we compare three ways to use each sealed self-probe
bank across limited anchor budgets: probe replay, anchor + probe replay, and
CPLUS.
Probe replay trains on frozen-model argmax labels. Anchor + probe replay also
receives the same real anchors and 512 self-probes as CPLUS; anchors are
repeated to balance their aggregate sampling weight against the probes.
CPLUS instead uses anchor and probe gradients to construct its preservation
mask. Each seed uses its own matched bank, and all methods share the update
budget. \Cref{fig:self-probe-role-anchor-scale} summarizes the matched
comparison across anchor budgets.

\subsection{Estimating and Blending Preservation Gradients}

Let \(\tau_r\) be the first adaptation step of epoch \(r\). At that epoch's
preservation refresh, CPLUS evaluates two losses at \(\theta_{\tau_r}\).
Here we show the refresh index explicitly; the main text omits it.
The past-example loss is
\[
\mathcal L_A(\theta_{\tau_r})
=
\frac{1}{A}
\sum_{(x,y)\in\mathcal P_A}
\ell(f_{\theta_{\tau_r}}(x),y),
\]
and the self-probe loss is
\[
\mathcal L_S(\theta_{\tau_r})
=
\frac{1}{N}
\sum_{(\widetilde x,\widetilde y)\in\mathcal S_A}
\ell(f_{\theta_{\tau_r}}(\widetilde x),\widetilde y).
\]
Their mean gradients are
\[
g_{A,r}=\nabla_\theta\mathcal L_A(\theta_{\tau_r}),
\qquad
g_{S,r}=\nabla_\theta\mathcal L_S(\theta_{\tau_r}).
\]
The first source is verified but sparse; the second is broader but inherits the
behavior of the base model. CPLUS does not require either gradient to be
an unbiased estimate of the unavailable complete-set gradient.

CPLUS combines the two sources before constructing an update mask:
\[
\widehat g_r
=
\lambda g_{A,r}
+
(1-\lambda)g_{S,r}.
\]
We use \(\lambda=\tfrac12\). This is a raw blend: we apply neither norm
alignment nor source-specific rescaling before addition. Equal coefficients do
not imply equal coordinate-wise influence because the source gradients may
differ in magnitude and structure. The norm-aligned alternative is evaluated
as an ablation rather than included in the default method.

\subsection{Masking the Optimizer Update}

Let \(\mathcal L_{\mathrm{new}}\) be the new-data objective. Adaptation steps
start at \(t=1\), with \(\theta_1=\theta_0\). At step \(t\), the optimizer
uses its complete internal state---including adaptive preconditioning and
momentum where applicable---to propose an update
\[
\delta_t
=
\operatorname{OPT}_t
\!\left(\nabla_\theta\mathcal L_{\mathrm{new}}(\theta_t)\right).
\]
Thus, without CPLUS, the parameters would change as
\(\theta_{t+1}=\theta_t+\delta_t\). CPLUS masks this proposed update rather
than the raw new-data gradient.

Let \(r(t)\) denote the epoch containing step \(t\), whose gradient was
refreshed at \(\tau_{r(t)}\). The instantaneous mask is
\[
m_{t,i}
=
\mathbb I
\!\left[
\widehat g_{r(t),i}\delta_{t,i}\leq0
\right].
\]
A coordinate is retained when its contribution to the loss linearization at
the latest refresh is nonpositive. Consequently,
\[
\left(\widehat g_{r(t)}\right)^\top
(m_t\odot\delta_t)
\leq0.
\]
This is CPL's coordinate-wise first-order decision, applied to the blended
preservation gradient. Following CPL's released implementation
\citep{yang2026cpl}, we refresh the preservation gradients once per epoch and
reuse their blend within that epoch. The update mask is recomputed at every
adaptation step.

\subsection{Temporal Smoothing}

Binary decisions may fluctuate when the preservation signal is estimated from
limited data. CPLUS maintains an exponential moving average of the
instantaneous masks,
\begin{align}
e_t
&=
\beta e_{t-1}+(1-\beta)m_t,
\qquad e_0=0,\\
\overline m_t
&=
\operatorname{clip}
\!\left(\frac{e_t}{1-\beta^t},0,1\right),
\end{align}
and applies
\[
\theta_{t+1}
=
\theta_t+\overline m_t\odot\delta_t.
\]
We use \(\beta=0.9\). Each effective coefficient remains in \([0,1]\), so the
soft mask can attenuate a proposed update but cannot amplify it or reverse its
direction. Values near one indicate repeated compatibility with the
preservation signal; values near zero indicate persistent conflict. This is an
operational interpretation of the mask, not a claim that individual
coordinates have an identifiable semantic meaning.

Temporal smoothing is a controlled relaxation rather than a strict per-step
safety guarantee. Because the EMA includes earlier decisions, a coordinate
rejected by the current binary mask can retain a nonzero coefficient. The
nonpositive inner-product condition above holds for \(m_t\odot\delta_t\)
with the cached preservation gradient, but not necessarily for the
EMA-smoothed update. After the refresh step, the cached gradient need not
equal the gradient at the current parameters. We evaluate this
stability--constraint trade-off using update-weighted mask diagnostics.

\subsection{Complete Procedure and Cost}

Self-probe generation is a one-time operation. During adaptation, CPLUS
refreshes the two mean preservation gradients at the beginning of each epoch,
forms one blended gradient, and reuses it for the adaptation updates in that
epoch. The model is optimized only on \(\mathcal D_{\mathrm{new}}\); the anchors and
self-probes influence adaptation exclusively through the mask.

Relative to ordinary fine-tuning, each refresh adds forward and backward
passes over \(A+N\) reference examples. CPLUS stores one blended preservation
gradient and one EMA state with the parameter shapes being masked; the two
source gradients are required only transiently during fusion. Algorithm
\Cref{alg:cplus} gives the complete procedure.

\section{CPLUS Pseudocode}
\label{app:cplus-pseudocode}

\begin{algorithm}[!htbp]
\caption{CPLUS}
\label{alg:cplus}
\begin{algorithmic}[1]
\Require Base-model parameters \(\theta_0\), new data \(\mathcal D_{\mathrm{new}}\), anchors \(\mathcal P_A\), \(N,K,T,\lambda,\beta\)
\State Freeze \(f_{\theta_0}\) and construct self-probes \(\mathcal S_A\)
\State Initialize \(t\gets1\), \(\theta_1\gets\theta_0\), and \(e_0\gets0\)
\For{epoch \(r=1,\ldots,R\)}
    \State \(\tau_r\gets t\)
    \State Compute mean gradient \(g_{A,r}\) from \(\mathcal P_A\) at \(\theta_{\tau_r}\)
    \State Compute mean gradient \(g_{S,r}\) from \(\mathcal S_A\) at \(\theta_{\tau_r}\)
    \State \(\widehat g_r\gets \lambda g_{A,r}+(1-\lambda)g_{S,r}\)
    \For{new-data minibatch \(B\subset\mathcal D_{\mathrm{new}}\)}
        \State \(\delta_t\gets \operatorname{OPT}_t(\nabla_\theta\mathcal L_B(\theta_t))\)
        \State \(m_t\gets \mathbb I[\widehat g_r\odot\delta_t\leq0]\)
        \State \(e_t\gets \beta e_{t-1}+(1-\beta)m_t\)
        \State \(\overline m_t\gets \operatorname{clip}(e_t/(1-\beta^t),0,1)\)
        \State \(\theta_{t+1}\gets \theta_t+\overline m_t\odot\delta_t\)
        \State \(t\gets t+1\)
    \EndFor
\EndFor
\State \Return \(\theta_t\)
\end{algorithmic}
\end{algorithm}

\section{Additional Analyses}
\FloatBarrier

The Qwen3 and Gemma evaluation grids use JAX/BF16 on TPUs. The Phi-3.5 Mini
out-of-family grid is a separate Torch/BF16 replication on two A100 GPUs with
FSDP and learning rate \texttt{1e-5}; the default learning rate is
\texttt{1e-6}. Each auxiliary diagnostic below keeps its execution setup fixed across
the compared methods; exact environments are recorded with the released
configurations.

\subsection{Model-Scale Analyses}
\label{app:model-scale}

\begin{figure}[!htbp]
    \centering
    \includegraphics[width=0.9\linewidth]{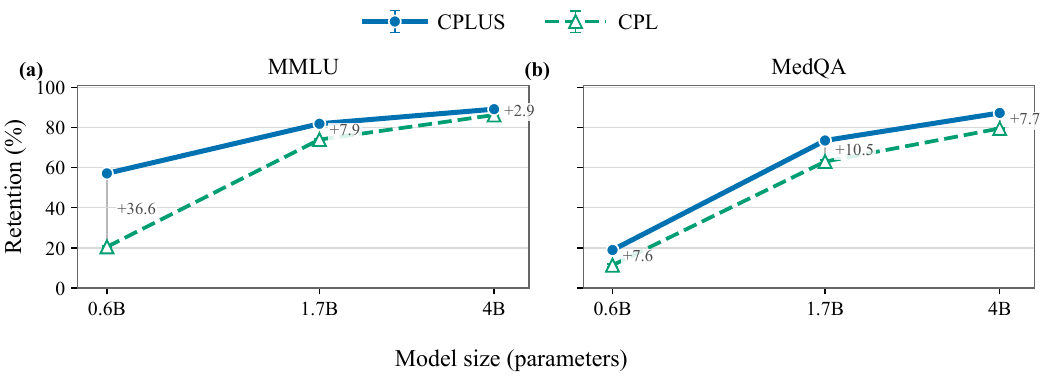}
    \caption{\textbf{Retention under limited anchor access across model
    scale.} Retention denotes Correct \(\rightarrow\) Correct.
    We compare access-matched CPL and CPLUS with four anchors on MMLU and
    MedQA. Labels report the absolute Retention
    improvement of CPLUS over CPL; points and error bars show means and
    standard errors. Every setting attains 100\% Acquisition (Incorrect
    \(\rightarrow\) Correct).}
    \label{fig:model-scale-headline}
\end{figure}

\Cref{fig:model-scale-headline} reports the access-matched comparison between
CPL and CPLUS with four anchors on MMLU and MedQA. Absolute
Retention increases with model size for both methods. We interpret the
comparison as a within-family trend.

\begin{figure}[!htbp]
    \centering
    \includegraphics[width=\textwidth]{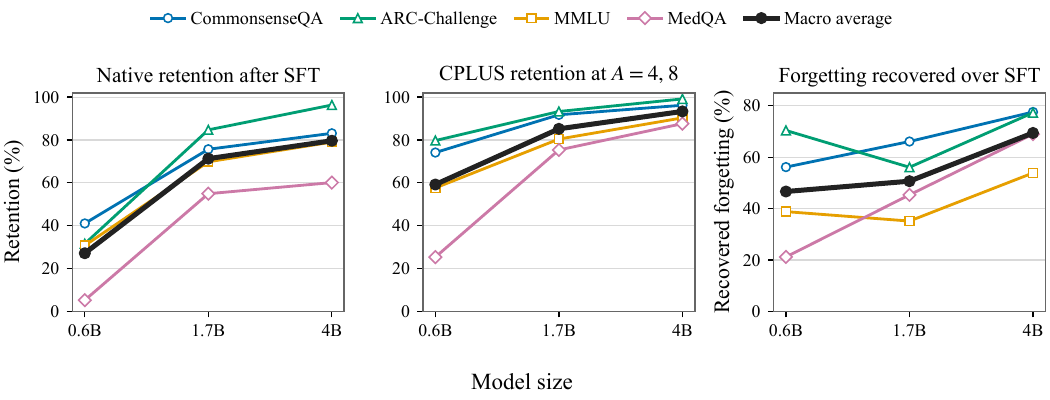}
    \caption{\textbf{Model-scale perspective.} Colored lines show individual
    benchmarks and the black line is their macro average. Retention denotes
    Correct \(\rightarrow\) Correct. \textbf{Left:}
    Retention after unconstrained fine-tuning, measuring each model's native
    vulnerability to forgetting. \textbf{Center:} CPLUS Retention averaged
    over \(A\in\{4,8\}\). \textbf{Right:} the fraction of SFT forgetting
    recovered by CPLUS, computed as
    \(100(R_{\mathrm{CPLUS}}-R_{\mathrm{SFT}})/(100-R_{\mathrm{SFT}})\),
    with CPLUS Retention averaged over the same two values of \(A\). Every
    CPLUS run in the fixed grid attains 100\% Acquisition (Incorrect
    \(\rightarrow\) Correct). Because each model induces its own mastered and
    new-data sets, the right panel measures the share of each model's
    available forgetting headroom that CPLUS recovers.}
    \label{fig:model-scale}
\end{figure}

The complementary four-benchmark perspective appears in
\Cref{fig:model-scale}. Its headroom-normalized panel accounts for the smaller
amount of native forgetting available at larger model sizes and shows the
fraction of that forgetting recovered by CPLUS.

\subsection{Hyperparameter Selection}
\label{app:hyperparameter-selection}

\begin{figure}[!htbp]
    \centering
    \includegraphics[width=\textwidth]{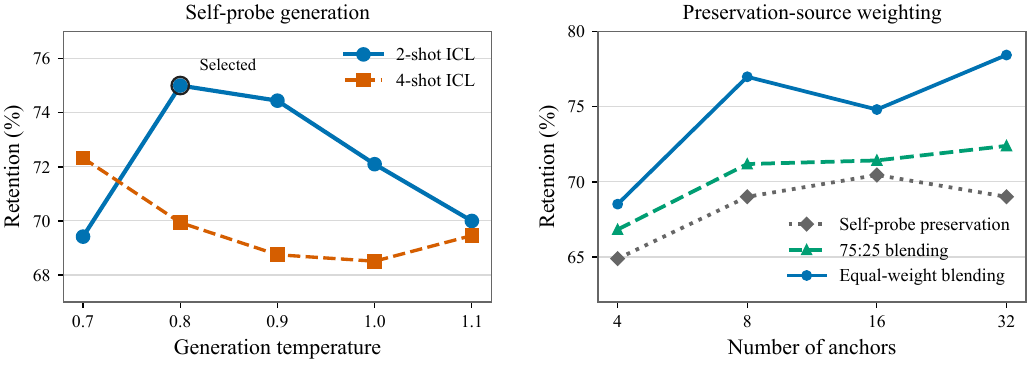}
    \caption{\textbf{Development-only hyperparameter searches.}
    Retention denotes Correct \(\rightarrow\) Correct.
    \textbf{Left:} Retention for two- and four-example in-context generation
    across sampling temperatures, evaluated with Qwen3-0.6B on CommonsenseQA
    at \(A=4\) and \(N=512\). The selected configuration uses two examples
    and \(T=0.8\). \textbf{Right:} Retention across
    \(A\in\{4,8,16,32\}\) for self-probe-only preservation, 75:25 blending,
    and equal-weight blending under a separate development protocol.
    Equal-weight blending is strongest at all four budgets. All runs achieve
    100\% Acquisition (Incorrect \(\rightarrow\) Correct).}
    \label{fig:hyperparameters}
\end{figure}

We select CPLUS hyperparameters using only the Qwen3-0.6B CommonsenseQA
development scenario. \Cref{fig:hyperparameters} summarizes the generation
and source-weight searches. With \(A=4\), two-example generation at
\(T=0.8\) achieves the highest observed Retention in the generation screen.
Equal-weight blending also outperforms self-probe-only preservation and 75:25
blending at every tested anchor budget.

The separate reference-reduction and smoothing screen in
\Cref{tab:ema-selection} favors mean reduction with \(\beta=0.9\), the default
used throughout the paper.

\begin{table}[!htbp]
    \centering
    \small
    \caption{\textbf{Reference reduction and mask smoothing.} Development
    results with \(A=64\), averaged over three runs.}
    \label{tab:ema-selection}
    \begin{tabular}{lccc}
        \toprule
        Reduction & EMA & Retention (\%) & Acquisition (\%) \\
        \midrule
        Mean & None & \(81.11 \pm 2.95\) & 100 \\
        Mean & 0.9 & \(\mathbf{82.32 \pm 0.64}\) & 100 \\
        CVaR & 0.9 & \(77.40 \pm 0.28\) & 100 \\
        \bottomrule
    \end{tabular}
\end{table}

\subsection{Epoch-Level Acquisition--Retention Dynamics}
\label{app:epoch-level-dynamics}

\begin{figure}[!htb]
    \centering
    \includegraphics[width=\textwidth]{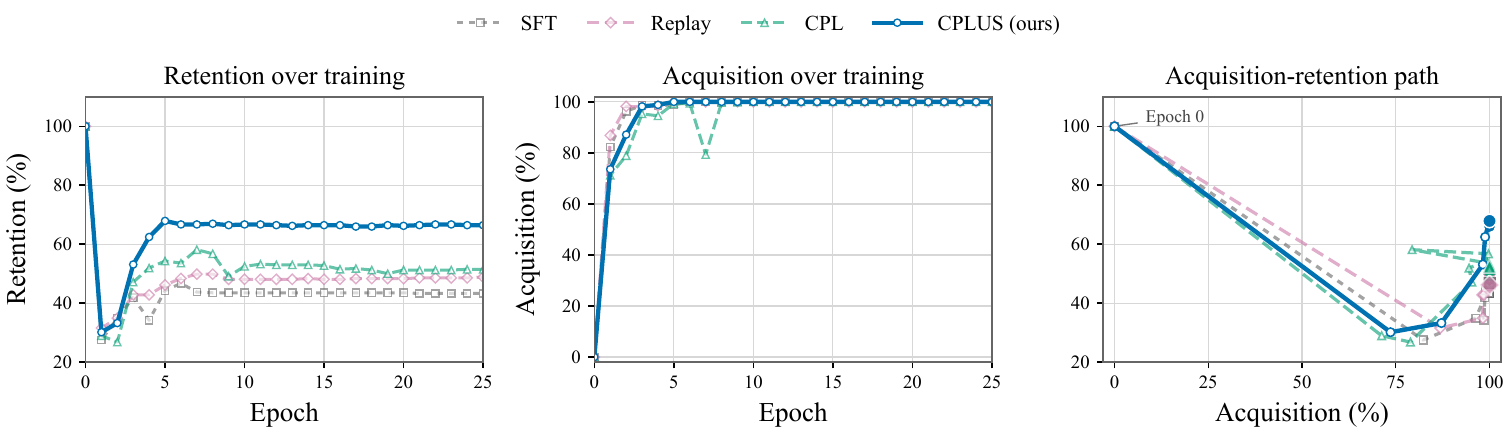}
    \caption{\textbf{Epoch-level Acquisition--Retention dynamics under limited
    anchor access.} We evaluate SFT, Replay, CPL, and CPLUS after every epoch
    over 25 epochs on Qwen3-0.6B and CommonsenseQA with \(A=8\) anchors and
    \texttt{lr=1e-6}. Left and center show Retention and Acquisition
    over training; right traces the corresponding Acquisition--Retention path.
    Both Retention panels use the same vertical scale, labeled on the left.
    Per-epoch uncertainty is not shown. Filled markers identify the first
    epoch reaching 100\% Acquisition.}
    \label{fig:epoch-level-dynamics}
\end{figure}

To examine how methods approach their final solutions, we evaluate Acquisition
and Retention after every epoch. This separate Torch/FP32 diagnostic uses one
A100 per run, rather than the JAX/BF16 TPU setup of the Qwen3 five-run grid;
its endpoints are not the endpoints of \Cref{fig:main-results}.
\Cref{fig:epoch-level-dynamics} shows Qwen3-0.6B on CommonsenseQA with
\(A=8\), \texttt{lr=1e-6}, and 25 epochs.

The trajectories reveal a sharp early drop in Retention as Acquisition rises
rapidly, followed by partial recovery during continued training. This pattern
is consistent with the previously reported \textit{stability gap}
\citep{delange2023continualevaluationlifelonglearning}. The methods differ in
both how quickly they learn the new examples and how much prior behavior
they recover: CPLUS reaches complete Acquisition by epoch 5 and maintains a
higher Retention plateau, whereas CPL first reaches complete Acquisition at
epoch 10.

These trajectories distinguish temporary forgetting from the forgetting that
remains at the end of adaptation. An early evaluation alone can miss
subsequent recovery, while final Retention alone can hide substantial
temporary forgetting. Reaching complete Acquisition also does not imply
equal recovery of prior behavior: methods that fully learn the adaptation
examples can still settle at different Retention levels.

\subsection{Anchor Composition Across Tasks}
\label{app:anchor-composition}

We test whether preservation anchors must be examples the base model already
answers correctly. We adapt Qwen3-0.6B to MedQA with CPLUS while fixing an
\(A=8\) budget of ground-truth-labeled CommonsenseQA anchors. Across rows in
\Cref{tab:anchor-composition}, only the composition of this small anchor set
changes; “correct” and “wrong” denote the base model's prediction before
adaptation. Retention and correction are measured on the complete held-out
base-correct and base-wrong CommonsenseQA subsets, respectively, excluding the
eight anchors.

The endpoint comparison shows the tradeoff: base-correct
anchors more strongly preserve established predictions, while base-wrong
anchors improve correction and overall accuracy.

\begin{table}[!htbp]
\centering
\small
\caption{Effect of anchor composition when adapting Qwen3-0.6B to MedQA with
\(A=8\) ground-truth-labeled CommonsenseQA anchors. All rows obtain 100\%
MedQA Acquisition and 100\% accuracy on the eight supplied anchors. Results
use one seed.}
\label{tab:anchor-composition}
\begin{tabular}{@{}ccccc@{}}
\toprule
\multicolumn{2}{c}{Anchor composition} & \multicolumn{3}{c}{CommonsenseQA (\%)} \\\\
\cmidrule(lr){1-2}\cmidrule(lr){3-5}
Base-correct & Base-wrong & Retention & Correction & Overall \\\\
\midrule
8 & 0 & 83.49 & 13.42 & 40.15 \\\\
6 & 2 & 81.34 & 21.24 & 44.16 \\\\
4 & 4 & 80.62 & 23.16 & 45.07 \\\\
2 & 6 & 80.38 & 20.80 & 43.52 \\\\
0 & 8 & 80.86 & 23.45 & 45.35 \\\\
\bottomrule
\end{tabular}
\end{table}

\subsection{Fixed Aggressive-Update Stress Test}
\label{app:fixed-aggressive-update}

We test optimization robustness under a fixed aggressive-update schedule.
Forgetting remains substantial at the default \texttt{lr=1e-6}, despite the
benefits of small finetuning learning rates \citep{marek2026forgetting}; the
larger-learning-rate schedule below produces more severe forgetting
(\Cref{fig:main-results,tab:fixed-aggressive-update}).

For this stress test, we apply one pre-specified
aggressive schedule to every method: AdamW with learning rate
\texttt{lr=1e-5} for exactly 10 epochs. We disable early stopping, epoch selection,
and Acquisition-based run filtering, and evaluate the final epoch. The model,
CommonsenseQA split, \(A=8\) access budget, optimizer, and seed are shared
across methods.

\Cref{tab:fixed-aggressive-update} shows that CPLUS matches SFT's 98.97\%
Acquisition while retaining 17.22 percentage points more of the base model's
correct predictions. Replay attains complete Acquisition but retains less, and
CPL reaches only 42.63\% Acquisition under this deliberately aggressive
schedule. The result therefore complements the Acquisition-controlled
comparisons: CPLUS continues to preserve more established behavior when every
method receives the same fixed optimization pressure.

\begin{table}[!htbp]
\centering
\small
\caption{Fixed aggressive-update stress test on Qwen3-0.6B and CommonsenseQA
with \(A=8\). Every method uses AdamW at \texttt{lr=1e-5} for exactly 10 epochs,
with no early stopping, epoch selection, or Acquisition filtering. Results
are from seed 0.}
\label{tab:fixed-aggressive-update}
\begin{tabular}{@{}lccc@{}}
\toprule
Method & Acquisition (\%) & Retention (\%) & Overall accuracy (\%) \\\\
\midrule
SFT & 98.97 & 21.77 & 69.53 \\\\
Replay & \textbf{100.00} & 12.68 & 66.70 \\\\
CPL & 42.63 & 9.57 & 30.02 \\\\
CPLUS & 98.97 & \textbf{39.00} & \textbf{76.09} \\\\
\bottomrule
\end{tabular}
\end{table}

\subsection{Matched Update Controls}
\label{app:matched-update-controls}

We test whether CPLUS retains more behavior merely because its mask reduces
the magnitude or density of the applied update. Norm-matched SFT preserves the
full SFT direction but rescales its update norm to the norm produced by CPLUS
at the same model state. The randomized-mask control permutes CPLUS's effective
mask coefficients within each parameter tensor at every step. This preserves
the exact per-tensor coefficient multiset and nonzero fraction, while
breaking the association between masked coordinates
and the preservation gradient.

This diagnostic uses the canonical Qwen3-0.6B CommonsenseQA setting with
\(A=8\), seed 0, the same sealed \(N=512\) self-probe bank, and the same
25-epoch schedule for all rows. \Cref{tab:matched-update-controls} shows that
neither matched update norm nor matched mask coefficients reproduce CPLUS's
Retention at the same Acquisition.

\begin{table}[!htbp]
\centering
\small
\caption{Matched update controls. Every row achieves 100\% Acquisition. The
randomized mask preserves CPLUS's exact within-tensor coefficient multiset,
while norm-matched SFT matches the norm of the update applied by CPLUS.}
\label{tab:matched-update-controls}
\begin{tabular}{@{}lccc@{}}
\toprule
Method & Matched property & Retention (\%) & Acquisition (\%) \\
\midrule
Norm-matched SFT & Applied update norm & 45.45 & 100 \\
Randomized CPLUS mask & Per-tensor mask coefficients & 47.13 & 100 \\
CPLUS & Probe-informed coordinate placement & \textbf{71.05} & 100 \\
\bottomrule
\end{tabular}
\end{table}

\subsection{Where the Preservation Mask Acts During Training}
\label{app:gradient-source-dynamics}

We examine how the preservation signals constrain proposed updates over
training in the canonical limited-anchor setting. CPL and CPLUS use separate,
matched Qwen3-0.6B CommonsenseQA runs with \(A=8\), seed 0, AdamW at
\texttt{lr=1e-6}, and 25 epochs; CPLUS uses its sealed \(N=512\) self-probe bank.
At seven pre-update states, a read-only audit previews the AdamW proposal on
the same deterministic 32-example adaptation subset and computes the
instantaneous gradient-sign mask for each method. For CPLUS, the diagnostic
uses the blended probe--anchor gradient before the mask's EMA smoothing.
The audit does not change training updates.

\begin{figure}[!htb]
    \centering
    \includegraphics[width=\linewidth]{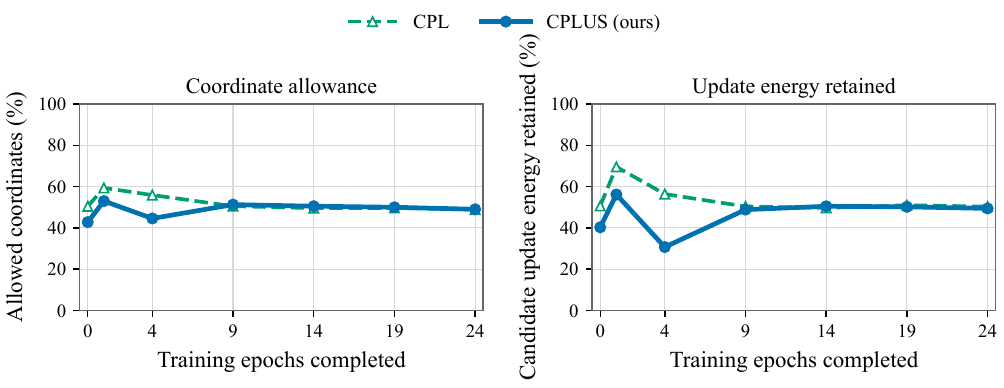}
    \caption{\textbf{Instantaneous mask geometry under limited anchor
    access.} CPL and CPLUS are audited at seven pre-update states in separate,
    matched Qwen3-0.6B CommonsenseQA runs with \(A=8\). The horizontal axis
    counts training epochs already completed. \textbf{Left:} fraction of
    coordinates admitted by the diagnostic gradient-sign mask.
    \textbf{Right:} fraction of each method's candidate AdamW update squared
    norm admitted by that mask. The CPLUS diagnostic uses its blended
    probe--anchor gradient before EMA smoothing. The audit is read-only;
    results use seed 0.}
    \label{fig:gradient-source-dynamics}
\end{figure}

\Cref{fig:gradient-source-dynamics} shows that CPLUS's instantaneous mask
admits fewer coordinates and less candidate-update energy in the early
states. From nine completed epochs onward, both methods admit approximately
half of the coordinates and half of their respective proposals' squared
update energy. These similar late-stage aggregate fractions do not account
for the final gap: both methods reach 100\% Acquisition, while CPLUS retains
71.05\% of initially correct behavior and CPL retains 50.48\%.

The placement control in \Cref{tab:matched-update-controls} isolates the
missing spatial information: permuting CPLUS's mask coefficients within each
parameter tensor preserves their exact multiset at every step but lowers
Retention to 47.13\% at the same 100\% Acquisition. Together, the temporal
audit and the placement control support the interpretation that
\emph{which coordinates are attenuated matters}, not only how many
coordinates or how much update energy the mask admits.

\subsection{Mask Behavior}
\label{app:mask-behavior}

We ask whether CPLUS's advantage comes from closer agreement with a full-data
update mask. We replay the same proposed updates across anchor budgets and
compare each method with its own full-data reference.

\begin{figure}[!htb]
    \centering
    \includegraphics[width=0.62\linewidth]{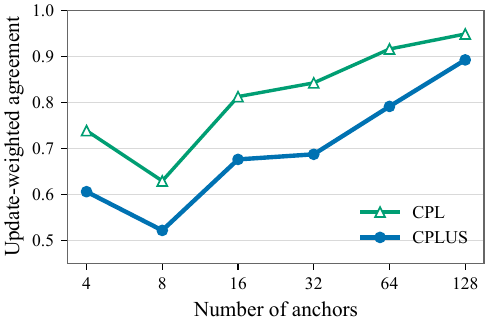}
    \caption{\textbf{CPLUS need not recover the full-data mask.} Each method
    is compared with its own full-data reference. CPL is closer at every
    budget despite lower Retention (Correct \(\rightarrow\) Correct), so mask
    similarity alone cannot explain CPLUS's advantage. Coordinate-level masks
    appear in \Cref{fig:mask-analysis-coordinates}.}
    \label{fig:mask-analysis}
\end{figure}

\begin{figure}[!htb]
    \centering
    \includegraphics[width=0.98\linewidth]{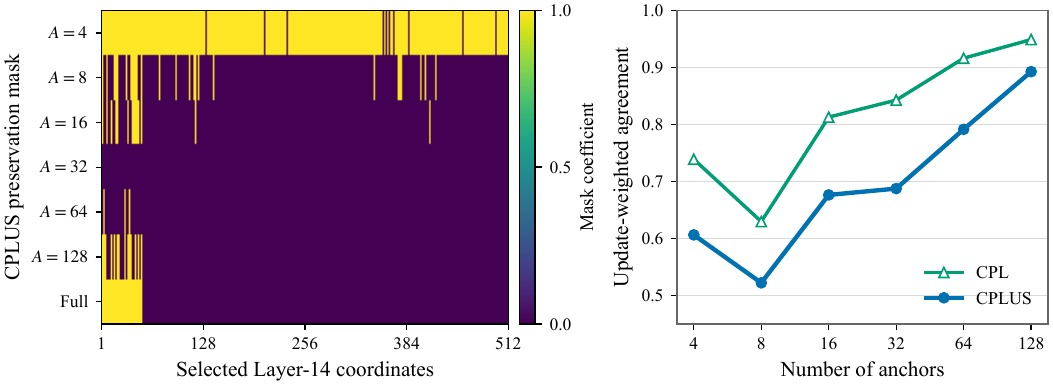}
    \caption{\textbf{Coordinate-level view of the mask audit.} We replay the
    same 678 proposed optimizer updates under controlled probe-side inputs.
    \textbf{Left:} each row shows the update-weighted mean CPLUS mask at one
    anchor budget; ``Full'' uses all 418 past examples. Columns are 512
    pre-registered coordinates from a middle-layer MLP down-projection,
    ordered identically in every row. Purple indicates a suppressed update
    coefficient (0) and yellow a retained coefficient (1). \textbf{Right:}
    each method is compared with its own full-data reference mask---CPL with
    full-data CPL and CPLUS with full-data CPLUS. Agreement is update-energy
    weighted, and 1 denotes identical masks.}
    \label{fig:mask-analysis-coordinates}
\end{figure}

\Cref{fig:mask-analysis} summarizes agreement with these references, while
\Cref{fig:mask-analysis-coordinates} shows the corresponding coordinate-level
masks. CPL more closely reconstructs its own
full-data mask at every budget. At \(A=128\), update-weighted agreement reaches
0.95 for CPL and 0.89 for CPLUS. Nevertheless, CPLUS achieves higher Retention
under limited access. Its improvement therefore does not arise from greater
agreement with its own full-data reference. Instead, blending the
self-probe and past-example gradients produces a distinct, temporally smoothed
update mask. Because this audit covers one model--benchmark pair, we treat it
as a diagnostic rather than a universal mechanistic explanation.

\subsection{Sequential Multi-Task Learning}
\label{app:sequential-multitask}

We update the model sequentially on CommonsenseQA, ARC-Challenge, MMLU,
and MedQA. All methods begin from the same post-CommonsenseQA checkpoint and
subsequently continue from their own preceding checkpoint. At each stage,
evaluation covers every task encountered so far. CPLUS maintains a separate
past-example set and self-probe bank for each preceding task and averages the
resulting task-level preservation gradients before masking the current-task
update.

\Cref{tab:sequential-stage-summary} reports the stage averages, while
\Cref{tab:sequential-final} breaks down accuracy by task after the final stage.

\begin{table}[!htbp]
    \centering
    \small
    \caption{\textbf{Average Task Accuracy during sequential multi-task
    learning.} Each entry is the unweighted mean of absolute accuracy over all
    tasks encountered through the indicated stage, averaged across three
    seeds. Task-level standard errors are reported in
    \Cref{tab:sequential-final} for the final stage and visualized throughout
    \Cref{fig:sequential-multitask}.}
    \label{tab:sequential-stage-summary}
    \begin{tabular}{@{}lccc@{}}
        \toprule
        Method & After ARC-C & After MMLU & After MedQA \\
        \midrule
        SFT & 66.81 & 72.20 & 61.21 \\
        Replay & 72.04 & 77.29 & 69.73 \\
        CPL & 75.18 & 70.77 & 69.65 \\
        \textbf{CPLUS} & \textbf{83.94} & \textbf{80.86} & \textbf{78.87} \\
        \bottomrule
    \end{tabular}
\end{table}

\begin{table}[!htbp]
    \centering
    \small
    \caption{\textbf{Accuracy after the final sequential-learning stage.}
    Following adaptation to MedQA, CPLUS achieves the highest accuracy on
    every task encountered in the sequence. The final Average Task Accuracy is
    the unweighted mean of the four absolute task accuracies. Values are means
    $\pm$ standard errors across three seeds; the Average column is computed
    from the task means.}
    \label{tab:sequential-final}
    \begin{tabular}{@{}lccccc@{}}
        \toprule
        Method & CSQA & ARC-C & MMLU & MedQA & Average \\
        \midrule
        SFT & $36.04\pm0.51$ & $53.78\pm0.49$ & $74.82\pm0.44$ &
        $80.19\pm0.43$ & 61.21 \\
        Replay & $55.29\pm0.49$ & $63.86\pm0.45$ & $79.59\pm0.41$ &
        $80.19\pm0.42$ & 69.73 \\
        CPL & $59.85\pm0.47$ & $62.60\pm0.45$ & $74.38\pm0.42$ &
        $81.77\pm0.40$ & 69.65 \\
        \textbf{CPLUS} & $\mathbf{69.34\pm0.46}$ &
        $\mathbf{78.27\pm0.44}$ & $\mathbf{83.43\pm0.44}$ &
        $\mathbf{84.45\pm0.42}$ & \textbf{78.87} \\
        \bottomrule
    \end{tabular}
\end{table}

\subsection{Base-Normalized Capability Scores}
\label{app:broader-capability}

\begin{figure}[!htb]
    \centering
    \includegraphics[width=0.72\linewidth]{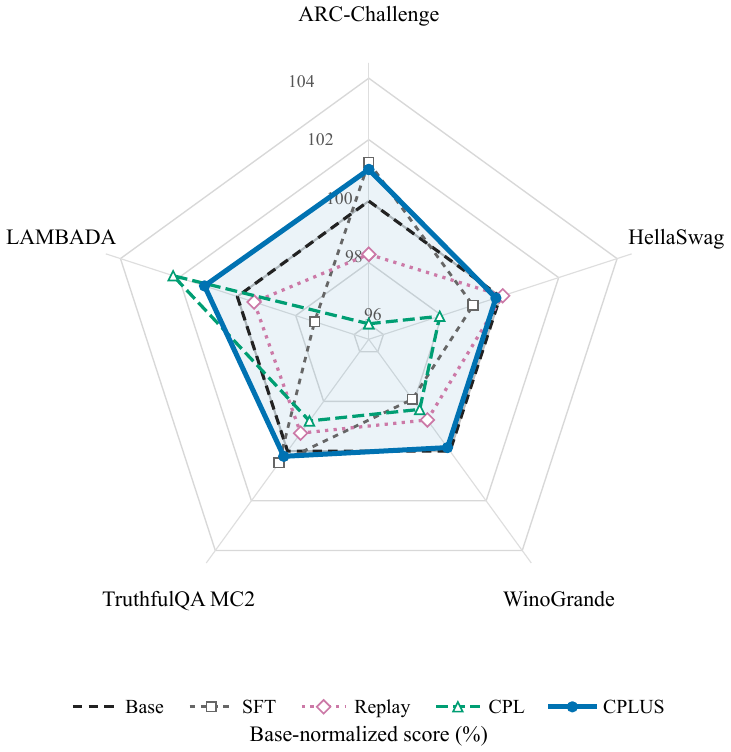}
    \caption{\textbf{Base-normalized capability scores after adaptation.}
    Each axis reports the post-adaptation score as a percentage of the
    corresponding base-model score. The dashed polygon marks the base-model
    level (100\%);
    values above 100 indicate improvement. Raw score changes are shown in
    \Cref{fig:problem-motivation}.}
    \label{fig:broader-capability-appendix}
\end{figure}

The main text reports per-benchmark changes after adaptation in the right panel
of \Cref{fig:problem-motivation}. For completeness,
\Cref{fig:broader-capability-appendix} shows the same evaluation normalized by
each benchmark's base-model score. CPLUS remains closest to the base capability
profile across the five benchmarks not used for adaptation.

\subsection{Layer-Wise Mask Audit}
\label{app:layerwise-mask-audit}

We extend the mask-behavior audit in \Cref{app:mask-behavior} across transformer
layers. The diagnostic uses Qwen3-0.6B on CommonsenseQA, the
sealed \(N=512\) self-probe bank, and a common sequence of 678 proposed
updates. Model weights are held fixed throughout the audit. For CPL, the
full-data reference uses all 418 past examples. For CPLUS, the method-matched
reference blends the same full past dataset with the controlled self-probe
inputs used throughout the audit.

For candidate mask trajectory \(m\), reference trajectory \(m^\star\), and
proposed updates \(\delta\), we report update-weighted agreement,
\[
\operatorname{Agreement}(m,m^\star)
=
1-
\frac{\sum_{t,i}\delta_{t,i}^{2}
\lvert m_{t,i}-m^\star_{t,i}\rvert}
{\sum_{t,i}\delta_{t,i}^{2}}.
\]
This emphasizes coordinates that materially contribute to the proposed
updates. The coordinate-level mask shown in
\Cref{fig:mask-analysis-coordinates} is the corresponding update-weighted temporal
mean. Its 512 coordinates are selected once using aggregate update energy and
then held fixed across all budgets and references.

\begin{figure}[!htb]
    \centering
    \includegraphics[width=\textwidth]{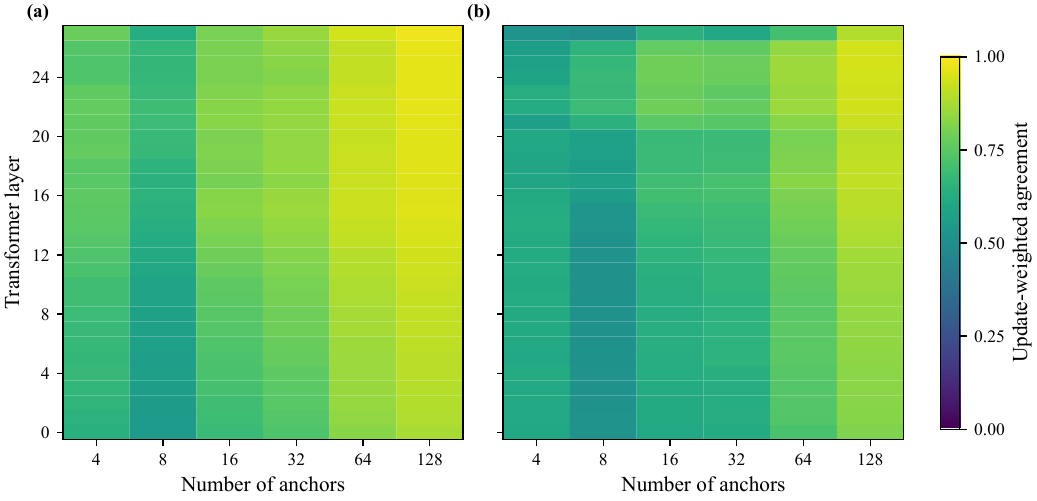}
    \caption{\textbf{Method-matched mask agreement across transformer depth.}
    Each cell reports update-weighted agreement for one layer and anchor
    budget. \textbf{(a)} CPL with \(A\) anchors compared with CPL using
    all 418 past examples. \textbf{(b)} CPLUS with \(A\) anchors compared
    with CPLUS using all 418 past examples under the same controlled probe
    inputs. The shared color scale ranges from zero to one.}
    \label{fig:mask-agreement-all-layers}
\end{figure}

\Cref{fig:mask-agreement-all-layers} shows the method-matched comparison
across all transformer layers. CPL remains closer to its full-data reference
than CPLUS across budgets and depth. Both methods become more stable as access
increases overall, but both exhibit a drop at \(A=8\); we therefore do not
claim monotonic convergence at the smallest budgets.

\begin{figure}[!htb]
    \centering
    \includegraphics[width=\textwidth]{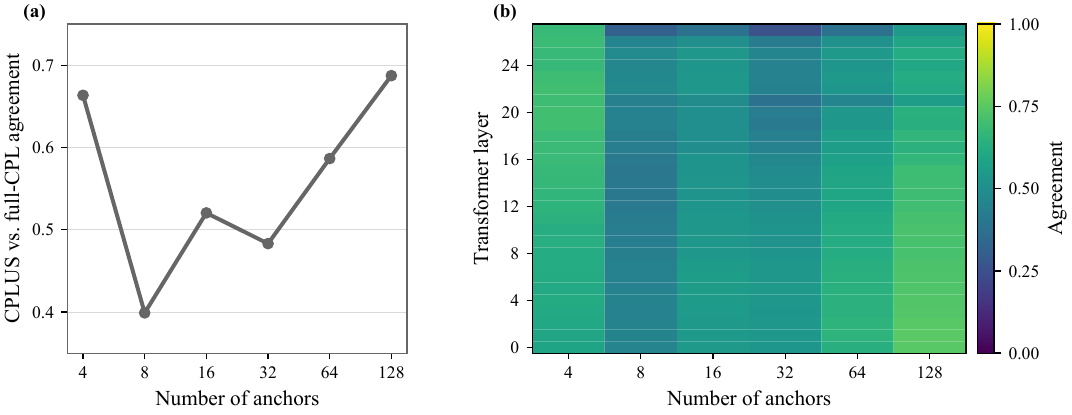}
    \caption{\textbf{Cross-reference comparison with full-data CPL.}
    CPLUS masks are compared directly with the mask constructed by CPL from
    all 418 past examples. \textbf{(a)} Agreement for the pre-registered
    Layer-14 MLP down-projection. \textbf{(b)} Agreement across all transformer
    layers. This deliberately mismatched reference measures how differently
    CPLUS suppresses updates; it is not an oracle target for CPLUS.}
    \label{fig:mask-cross-reference}
\end{figure}

For completeness, \Cref{fig:mask-cross-reference} directly compares
CPLUS with the all-real CPL reference. This is a cross-method diagnostic, not
the target that CPLUS is intended to reconstruct. The low agreement shows
that the two methods suppress different updates. Whether that difference
preserves broader function is evaluated behaviorally rather than inferred
from mask similarity.

\subsection{Evaluation and Compute Details}
\label{app:recurring-compute}

In the primary grid, Acquisition covers the initially incorrect adaptation
examples; Retention covers all initially correct examples, including anchors.
The anchor-composition diagnostic excludes its anchors
(\Cref{app:anchor-composition}); external-task averages exclude adaptation
and anchor benchmarks.

\Cref{fig:retention-compute} compares the three Qwen3 scales on CommonsenseQA
at \(A=8\). CPLUS uses \(N=32\) rather than the default 512 to reduce compute.
Frozen distillation matches frozen-teacher output distributions during
adaptation \citep{hinton2015distilling}. OGD-GTL projects new-data gradients
orthogonally to a basis of anchor ground-truth-logit gradients
\citep{farajtabar2020ogd}.

Training FLOPs are computed using the leading dense-transformer approximations \(2PL\)
per forward pass and \(6PL\) per forward-plus-backward pass, where \(P\) is
model size and \(L\) is padded sequence length. Repeated preservation
passes are included; evaluation and one-time preparation, including probe
generation and SSR refinement, are excluded. FLOPs describe arithmetic cost,
not elapsed time.

\section{Limitations}
\label{app:limitations}

Self-probes record the frozen model's behavior, not independently verified
truth. If that behavior is biased or poorly covered by the probes, the
resulting preservation signal can limit useful adaptation updates. Our scale
results characterize the tested models, and the sequential results concern
the tested task order. These findings do not establish preservation of every
capability or robustness to arbitrary task sequences.

\section{Extended Related Work}
\label{app:extended-related-work}

\paragraph{Gradient- and parameter-level preservation.}
Continual learning methods often mitigate interference by constraining updates
according to information retained from earlier tasks. Elastic Weight
Consolidation penalizes changes to parameters estimated to be important for
prior tasks \citep{kirkpatrick2017ewc}, while GEM and A-GEM use gradients from
episodic memory to prevent new-task updates from increasing previous-task loss
\citep{lopezpaz2017gem,chaudhry2019agem}. OGD instead projects updates away
from directions that alter previously learned outputs
\citep{farajtabar2020ogd}. More recently, CPL decomposes gradient interaction
coordinate-wise and suppresses updates predicted to conflict with previously
correct behavior \citep{yang2026cpl}. CPL also evaluates partial access to
these samples. With limited data, its preservation gradient can miss behavior
outside the retained examples. CPLUS complements the available
samples with self-probes from the frozen base model and blends the two
signals before masking the adaptation update.

\paragraph{Model-generated supervision.}
Recent work shows that a language model's own predictive distribution can
support continued adaptation~\citep{cho2026correctreasoningpathsvisit}. Self-generated replay uses samples from the
model's prior distribution as replay data and can substantially reduce
forgetting when sufficient capacity remains \citep{marek2026forgetting}.
Self-Distillation Fine-Tuning conditions the model on demonstrations and uses
its on-policy predictions as supervision for acquiring new skills
\citep{shenfeld2026selfdistillation}. RL's Razor attributes the lower
forgetting of on-policy reinforcement learning to its preference for solutions
that remain closer to the original policy \citep{shenfeld2025razor}. CPLUS
shares the principle that the base model contains information worth
preserving, but self-probes are neither replay samples nor targets for the
new behavior. They are used only to construct the gradient that guides
the preservation mask.

\paragraph{Capacity, plasticity, and model scale.}
Forgetting depends on the model's pretraining state, remaining capacity, and
ability to remain plastic. Knowledge entropy has been observed to decline
during language-model pretraining alongside weaker knowledge acquisition and
retention \citep{kim2025knowledgeentropy}, while forgetting can persist when a
model approaches capacity saturation \citep{marek2026forgetting}. Loss of
plasticity is distinct from forgetting: a model may preserve earlier behavior
yet become progressively less able to acquire new information
\citep{dohare2024plasticity, cho2026treeconceptsinterpretablecontinual}. When memory is abundant, the principal difficulty
can consequently shift from retention toward acquisition
\citep{cho2025forget}. These findings motivate measuring both quantities and
separating native robustness, absolute CPLUS performance, and the incremental
gain over matched baselines.

\paragraph{Evaluation and efficiency.}
Tuning and evaluating a method within the same continual learning scenario can
overestimate its ability to generalize \citep{cha2025hyperparameters}. We adopt
the corresponding two-phase principle by selecting hyperparameters on one
development setting and applying the self-probe and masking settings unchanged across
the reported model sizes, benchmarks, past-data budgets, and random seeds.
Under explicit compute constraints, simple memory-sampling baselines can match
or outperform more elaborate methods \citep{prabhu2023budgeted}, and some
continual learning systems can consume more computation than retraining from
scratch \citep{harun2023efficient}. These findings motivate our comparison with
replay and access-matched CPL, and our use of a simple global gradient blend.

\end{document}